%% file: root.tex
\documentclass[letterpaper, 10pt, conference]{ieeeconf}      

\IEEEoverridecommandlockouts                              
\usepackage{graphicx} 
\usepackage{amsmath} 
\usepackage{amsfonts}
\usepackage{mathtools} 

\usepackage{multirow}
\usepackage{tabularx}
\usepackage{caption}
\usepackage{subcaption}
\usepackage{hyperref}
\usepackage{float}
\usepackage{eucal} 
\usepackage{siunitx}
\usepackage{url}

\usepackage{algorithm} 
\usepackage[noend]{algpseudocode}

\input{tex/00_symbols.tex}
\title{\LARGE \bf PennSyn2Real: Training Object Recognition Models\\ without Human Labeling}

\author{Ty Nguyen$^{1*}$, Ian D. Miller$^{1*}$, Avi Cohen$^{1}$, Dinesh Thakur$^{1}$, Shashank Prasad$^{2}$,\\
Camillo J. Taylor$^{1}$, Pratik Chaudhari$^{1}$, Vijay Kumar$^{1}$
\thanks{We gratefully acknowledge the support from ARL Grant DCIST CRA W911NF-17-2-0181, NSF Grant CNS-1521617,
ARO Grant W911NF-13-1-0350, ONR Grants N00014-20-1-2822 and
ONR grant N00014-20-S-B001, Qualcomm Research, C-BRIC, a Semiconductor Research Corporation Joint University Microelectronics Program cosponsored by DARPA, and NVIDIA. Ian Miller acknowledges the support of a NASA Space Technology Research Fellowship.}
\thanks{$^{1}$The authors are with the
GRASP Lab, University of Pennsylvania, Philadelphia, PA 19104 USA. {
        {\tt\footnotesize email: \{tynguyen, iandm, avrahamc, tdinesh, pratikac, cjtaylor,  kumar\}}@seas.upenn.edu}
}%
\thanks{$^{2}$ The authors are with Amazon. {\tt\footnotesize email: shashankdp@yahoo.com}}
\thanks{$^{*}$ Authors contributed equally}
}

\begin{document}

\maketitle
\thispagestyle{empty}
\pagestyle{empty}

\begin{abstract}
Scalable training data generation is a critical problem in deep learning. We propose PennSyn2Real - a photo-realistic synthetic dataset consisting of more than 100,000 4K images of more than 20 types of micro aerial vehicles (MAVs). The dataset can be used to generate arbitrary numbers of training images for high-level computer vision tasks such as MAV detection and classification. Our data generation framework bootstraps chroma-keying, a mature cinematography technique with a motion tracking system, providing artifact-free and curated annotated images where object orientations and lighting are controlled. This framework is easy to set up and can be applied to a broad range of objects, reducing the gap between synthetic and real-world data. 
We show that synthetic data generated using this framework can be directly used to train CNN models for common object recognition tasks such as  detection and  segmentation.  We demonstrate competitive performance in comparison with training using only real images. Furthermore, bootstrapping the generated synthetic data in few-shot learning can significantly improve the overall performance, reducing the number of required training data samples to achieve the desired accuracy.

\end{abstract}


\input{tex/01_introduction}

\input{tex/02_related_work}
\input{tex/03_data_collection_process}
\input{tex/04_MAV_dataset}

\input{tex/05_experiments}
\input{tex/06_conclusions}
\bibliographystyle{IEEEtran}
\bibliography{root}

\end{document}

%% file: tex/01_introduction.tex
\section{Introduction}
Safety-critical operation in applications such as autonomous driving and surveillance requires accurate predictions from perception systems. To achieve this level of accuracy, deep learning models often require a surprisingly large amount of training data.  One successful solution to this problem is transfer learning which initially trains a deep learning model on a large dataset such as ImageNet~\cite{imagenet_cvpr09} before finetuning the model on a smaller dataset tailored to the downstream task. In addition to ImageNet, other image datasets such as COCO~\cite{lin2014microsoft} and PASCAL VOC~\cite{everingham2010pascal} have been used successfully in transfer learning. However, these data may not be applicable for a specific task or object such as drone detection in a cluttered environment due to the covariate shift. On the other hand, tailoring data for a specific task with human labeling is time and cost prohibitive. Therefore, there is a high motivation for developing more scalable data generation techniques to cope with various downstream tasks more efficiently.



\begin{figure}[t]
    \centering
    \includegraphics[width=0.98\linewidth]{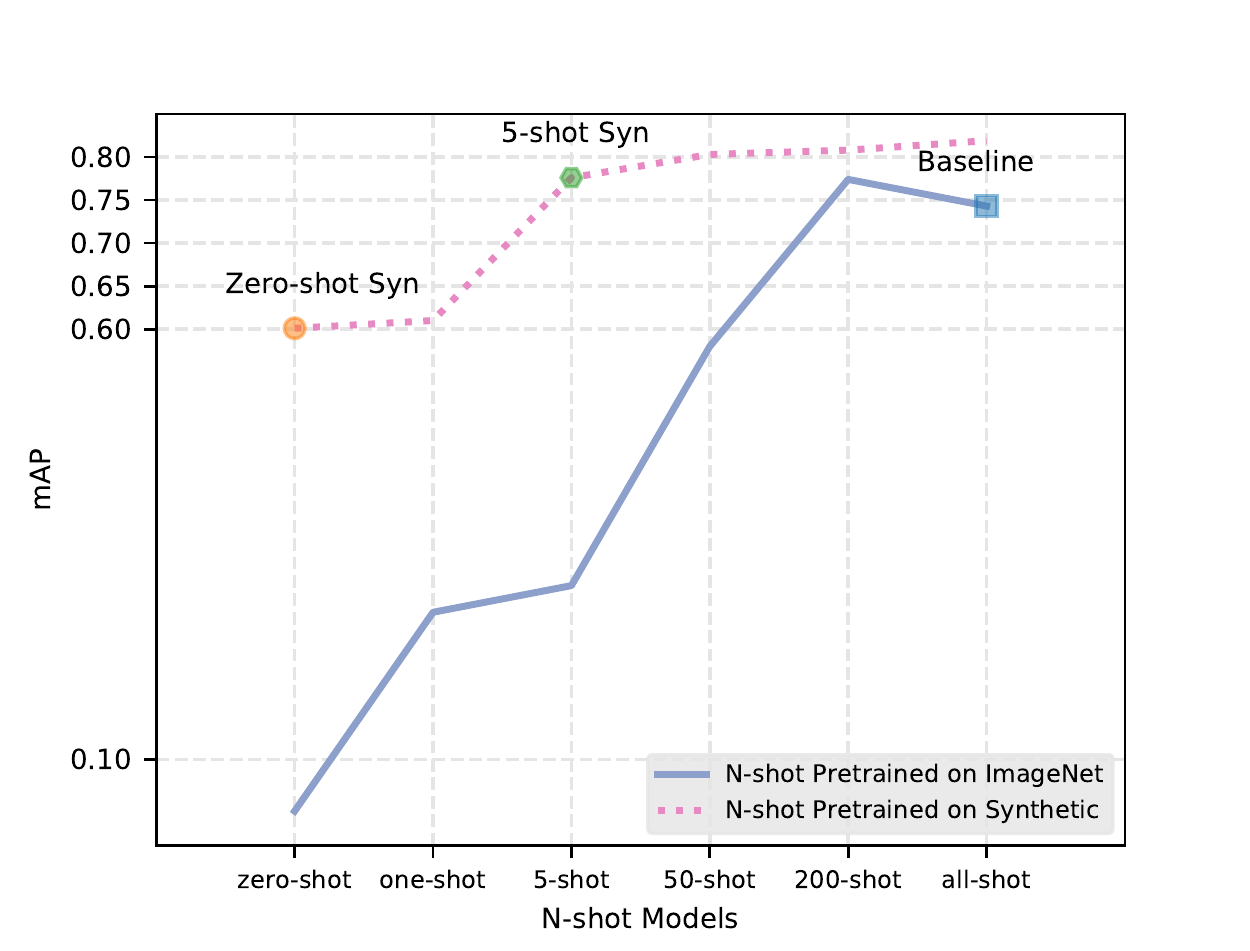}
    \caption{Quantitative performance comparison between Yolov3 models pretrained on ImageNet and those pretrained on the synthetic multi-MAV dataset when finetuning using different numbers of real training samples per object class. The baseline is Yolov3 pretrained on ImageNet before finetuning using all real training data. Models pretrained on the synthetic data perform better than their counterparts.}
    \label{fig:few-shot_detection_performance}
\end{figure}

There is large interest in solving computer vision problems using synthetic data thanks to its promise of dramatically simplifying the training process. On top of the arbitrary quantities of data that can be generated, synthesized data have perfectly accurate labels that human-labeling can not achieve. Indeed, authors in ~\cite{gaidon2016virtual, de2017procedural, nowruzi2019much} have demonstrated that synthetic datasets are a cost-effective alternative and supplement to manually labeled data and have good transferability.



The primary challenge when fabricating datasets is the sufficiently accurate reproduction of real-world data.  Synthetic data generated using game engines such as Unity are not yet ideal for training state-of-the-art (SoTA) DNNs due to the limited syn-to-real transferability.  The key reason for this issue, as Geirhos et al.~\cite{GeirhosRMBWB19} show empirically, is that SoTA DNNs often fail to utilize the shape of an object and instead heavily rely on textures. Despite recent advances in synthetic-to-real domain adaptation and refinement, there is still a gap in syn-to-real transferability.  

An alternative method for generating synthetic data with realistic textures is to superimpose real images onto real background images. A popular approach is to use human-labeled object's masks to extract objects from the original image and superimpose them onto any arbitrary background images. However, this approach results in boundary artifacts due to the inaccuracy of human labeling. 

To minimize these artifacts, we exploit the chroma-keying technique widely used in the filming industry to obtain high-quality object images. We then couple this technique with a motion capture system, as shown in Fig.~\ref{fig:system_settings}, to track the camera and object during video capture. This additional step provides additional labeled poses to each object image. While this work focus on object detection and tracking problems, the orientation information can be useful for other vision problems such as pose estimation.  It can also be used for guaranteeing uniform samples over the object pose when creating training data.


The popularity and accessibility of MAVs for commercial and recreational use have surged.  While being highly useful in various applications~\cite{miller2020mine, chen2020sloam, tang2020multi, saldana2018modquad}, MAVs can also pose security threats to sensitive areas such as airports and military bases. As we consider swarms of MAVs, it is necessary for MAVs to detect, classify and track each other. In all these applications, it is crucial to develop fine-grained analysis methods that are capable of fast and accurate detecting different models.  These tasks are challenging due to the various types of robots and cluttered environments in which MAVs operate.  We, therefore, introduce PennSyn2Real, a synthetic dataset consisting of more than 100,000 4K images of more than 20 types including our customized MAVs and commercial MAVs such as DJI Mavic Pro, Skydio 2, and Autel.  Furthermore, we demonstrate the transferability of our dataset by using synthetic data for training a CNN to detect and classify MAVs in both indoor and outdoor, cluttered environments.  
In short, this work centers around the following contributions:
\begin{enumerate}
    \item A large-scale MAV dataset with high variance in viewpoints for multi-MAV detection and recognition. 
    \item A data generation framework for scalable image data generation for training CNNs. 
    \item Validation of our framework and dataset by using synthetic data for training CNNs for object recognition tasks.
    \item Demonstration that this framework can be used to generate an alternative to ImageNet in the few-shot and transfer learning regimes.  
\end{enumerate}



%% file: tex/02_related_work.tex
\section{Related Works}
\subsection{Synthetic Datasets}


In general, many problems of modern AI come from the insufficiency of training data: either the available datasets are small, or generating labeled data is cost-prohibitive~\cite{nikolenko2019synthetic}. Thus, it is not surprising to see many interests in developing synthetic data generation techniques. Butler et al. (MPI-Sintel)~\cite{butler2012naturalistic}, Kaneva et al.~\cite{kaneva2011evaluation}, Baker et al. (Middlebury)~\cite{baker2011database, scharstein2014high} are among modern works on using animated movies to generate massive amounts of labeled images to train CNNs for low-level computer vision problems: optical flow estimation, and stereo correspondence estimation. They show that animated movies provide a feasible way to obtain large datasets with ground truth which would be otherwise hard and expensive to do. Dosovitskiy et al.~\cite{dosovitskiy2015flownet} contribute to this field by providing a large synthetic dataset called Flying Chairs from a public database of 3D chair models, adding them on top of real backgrounds to train a CNN-based optical flow estimation model.  

There have been many works that utilize synthetic image data to recognize everyday objects such as retail items, food, and furniture. For example, ShapeNet developed by Chang et al.~\cite{chang2015shapenet} provides millions of 3D CAD models classified into $3,135$ categories that can be used to synthesize images. This work, however, focuses on the shape of objects without realistic texture.  Additionally, the method requires the existence of 3D models for classes of interest.

Recent advances in photorealistic computer graphics platforms have benefited high-level computer vision problems such as object detection and semantic segmentation.  For example, Richter et al.~\cite{richter2017playing} obtain datasets from the Grand Theft Auto V video game and focus on semantic segmentation. While getting pixel-wise labels for segmentation is still done manually, they can cut the labeling cost by capturing the communication between the game and the graphics hardware. Thanks to the development in game engines such as Unreal Engine 4 that enables full control of camera location and field of view, it is latter feasible to get object notations from animated scenes without human labeling. The SYNTHIA dataset~\cite{ros2016synthia, hernandez2017slanted}, a synthetic dataset generated using game engines, for example, is shown to improve CNN's performance on a scene segmentation task. The virtual KITTI dataset~\cite{gaidon2016virtual, cabon2020virtual}, another game engine-based synthetic dataset, has a broader scope. Their system can generate training data for a wider range of applications, including semantic segmentation and depth estimation.  In SceneNet~\cite{mccormac2017scenenet}, authors demonstrate that
a semantic segmentation CNN pretrained from scratch on purely synthetic data can improve over such a model pretrained on ImageNet. All these datasets, however, focus on driving scenarios, and because they are generated using game engines they still suffer from artifact problems.


Unlike those datasets, PennSyn2Real is unique in that it combines the realistic appearance of an actual object and real background to generate arbitrary amounts of high-quality image data with realistic textures and minimal artifacts. Moreover, we use the motion capture system to control the variance in the object's orientations. PennSyn2Real's methodology can be useful for different applications from object recognition to pose estimation. Finally, we use only images of the object of interest, requiring no detailed CAD modelling of the object of interest or larger environments.

Recent literature has seen synthetic data creation in a more complex way by placing everyday objects in real surroundings. Georgakis et al.~\cite{georgakis2017synthesizing} propose procedures to superimpose synthetic objects onto real indoor backgrounds. Since objects are extracted from human-labeled datasets, there is little control over the variance in object's poses as well as lighting conditions and other factors, not to mention the unavoidable boundary artifacts caused by imperfect labeling. 

While whether photorealism is important for synthetic image data to transfer well to real images in high-level computer vision tasks is still debatable, Saleh et al.~\cite{sadat2018effective} argue that not all classes in a semantic segmentation problem are equally suitable for synthetic data. Indeed, objects such as people, cars, bikes, and others defined in the terminology of~\cite{andriluka20142d} are well suitable for object detectors that utilize shape a lot but suffer from the syn-to-real transfer for segmentation that largely relies on the texture.  Furthermore, Movshovitz-Attias et al.~\cite{movshovitz2016useful} argue that photorealistic rendering does indeed help. These arguments highlight the advantage of using our framework over other synthetic data generation approaches since it provides realistic textures for the real foreground and background images.

Wei et al.~\cite{wei2019rpc} and Follman et al.~\cite{follmann2018mvtec} propose a similar approach to generate synthetic data for training object recognition CNNs. However, they focus on retail objects in indoor, controlled settings. Furthermore, they do not label the object's pose in each image. 


\subsection{MAV datasets}
The literature has seen increasing interest in vision-based MAV detection and tracking. Unlu~\cite{unlu2019deep} and Wyder~\cite{wyder2019autonomous} develop deep learning-based approaches to detect MAVs using one or multiple cameras. Wyder et al.~\cite{wyder2019autonomous} also provide $58,647$ labelled MAV images. However, they are mostly indoor and specific to one single target MAV.  Jing et al.~\cite{li2016multi} introduce $70,000$ images of only a single type of MAV.  In addition to the limited variety of MAVs, these datasets are tailored specifically to specific environments, making them not suitable for adoption in other settings which involve environmental and object changes. By contrast, PennSyn2Real features more than 100,000 4K images of more than 20 types of different MAVs which can be used to generate an endless number of training images using different background images to better fit different environments. 

%% file: tex/03_data_collection_process.tex
\section{Data Generation Process}
\subsection{Object Image Collection}
\begin{figure}[t]
    \centering
    \includegraphics[width=0.8\linewidth]{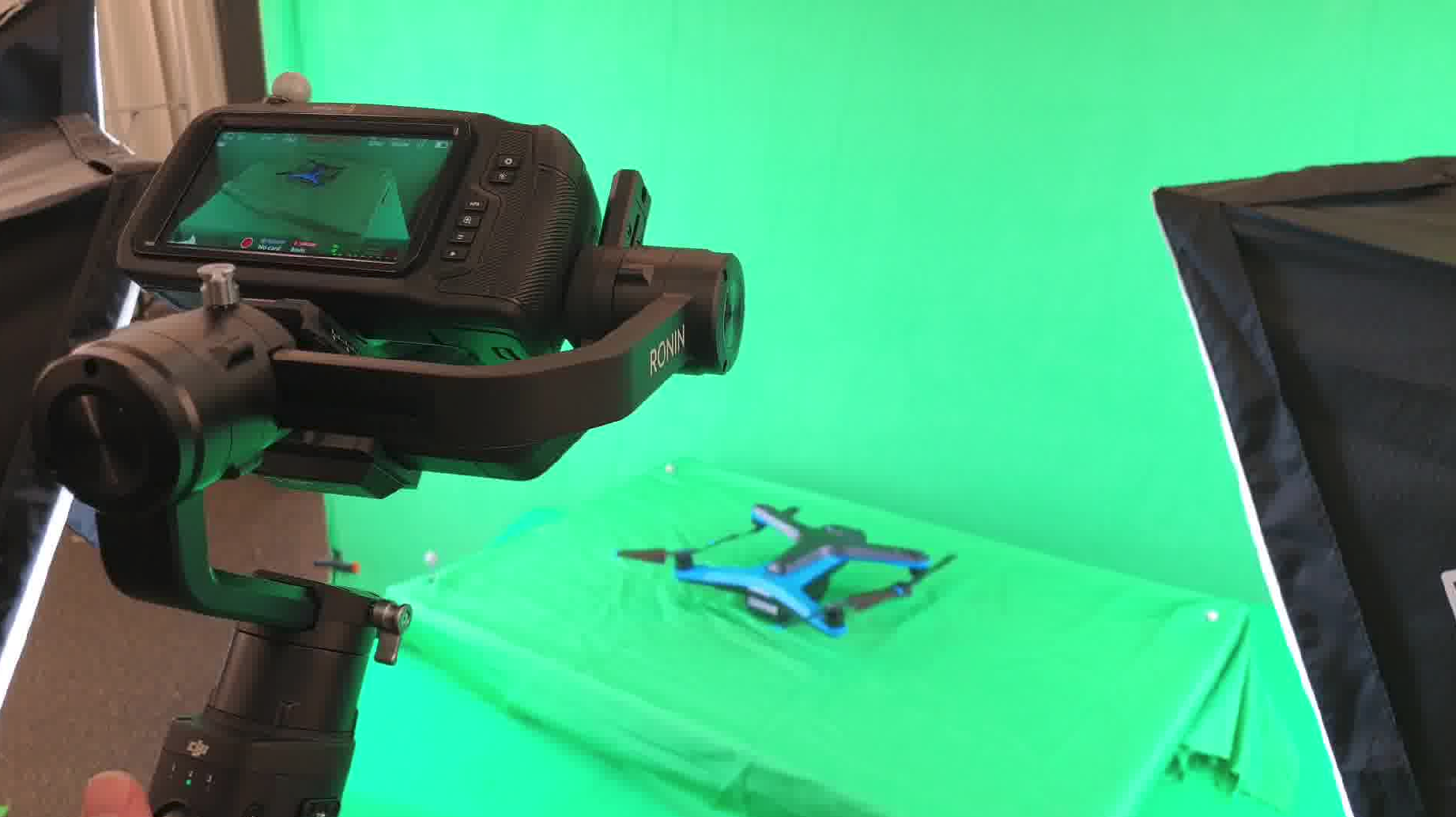}
    \caption{System settings for our object image collection. Object is put on a turntable covered by a green screen. Both the camera and turntable are tracked in a VICON system using markers}
    \label{fig:system_settings}
\end{figure}

\begin{figure}
    \centering
    \includegraphics[width=0.8\linewidth]{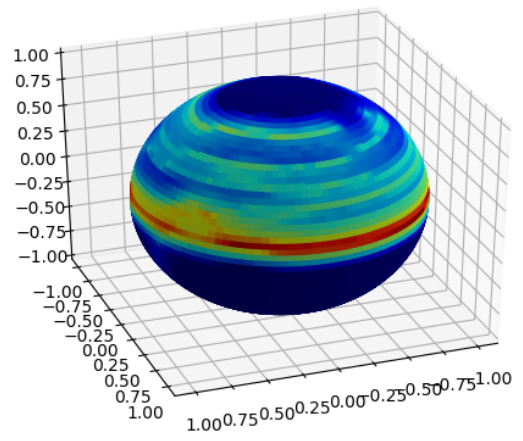}
    \caption{Distribution of viewing angles for one of the robots in the dataset.  The heatmap denotes the density of samples from the viewing direction determined by a ray from the origin through that point on the sphere.}
    \label{fig:rot_dist}
\end{figure}

Our approach focuses on object instances and their superimposition into real scenes at different positions, orientations, and scales, while reducing the discrepancy in lighting conditions and minimizing boundary artifacts.
We start by collecting high-quality 4K object images. Our setup, as shown in Fig.~\ref{fig:system_settings}, consists of a moving camera to capture 4K HDR videos of objects in front of a green screen. The background and object are separately lit by lightboxes to ensure uniform lighting for better chroma-keying.  The object is placed on a powered rotating turntable; the camera on a stabilizing gimbal.  In this way, with the gimbal adjusting the altitude and the turntable adjusting the azimuth, we control $2$ degrees of freedom of rotation. The final degree of freedom about the camera axis can be controlled by rotating the object image in post-processing, and the translational degrees of freedom by translation and scaling of the object image.  We show the distribution of viewing angles over this space in Fig.~\ref{fig:rot_dist}, and note that we well-cover the top-half of this distribution in this dataset, with exceptions being the bottom and very top of the object.  We additionally attach markers to the camera and turntable and track both in a motion capture system to additionally generate pose labels.  Motion capture and camera synchronization is performed in post-processing by aligning a rapid rotation of the camera made at the beginning of each video.

We use DaVinci Resolve, a commercial video editing software, to remove color bleed from the green screen and generate the chroma-keyed segmentation.

\subsection{Synthetic Data Generation}
\begin{figure}
    \centering
    \includegraphics[width=1.0\linewidth]{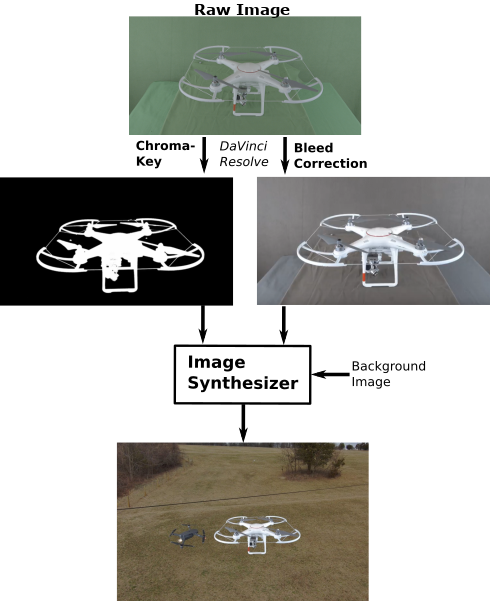}
    \caption{Pipeline for synthetic image generation.}
    \label{fig:synth_pipeline}
\end{figure}

\begin{figure}[t]
    \centering
    \includegraphics[width=0.9\linewidth]{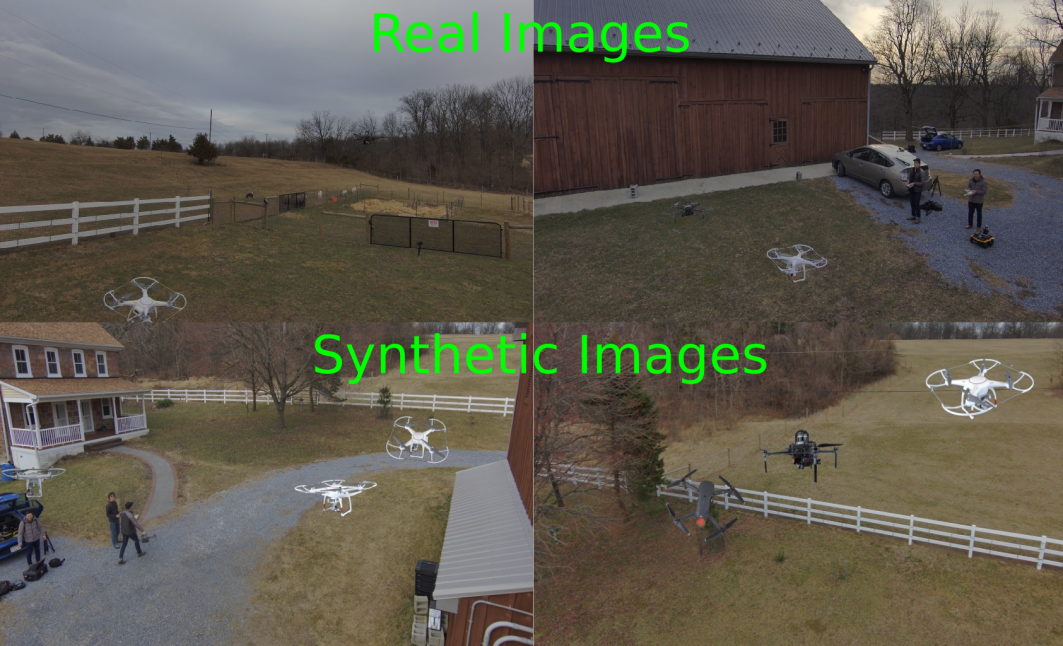}
    \caption{An example of the synthetic images in comparison with real images under the same surrounding environment}
    \label{fig:example_multi_robot}
\end{figure}

The data sample generation process is shown in Fig.~\ref{fig:synth_pipeline}.  First, we collect background images from the internet or by recording the surroundings under which we want to test a trained model. Next, we randomly choose an image of the object of interest and mask it out. To determine the object scale, one can either use the depth value of the chosen pose or randomly select a value. Finally, the object is blended into the background at the chosen pose with an appropriate brightness adjustment. An example of a synthetic image in comparison with a real image under the same scene is shown in Fig.~\ref{fig:example_multi_robot}.   

In our experiments, we randomly select a scale from $0.1$ to $0.3$ to approximate the scale of objects appearing in our real datasets. To mitigate the effects of illumination and contrast changes when blending the object with the background image, one can use Seamless Image Cloning technique~\cite{tanaka2012seamless} or simply Gaussian filters.  We also apply the inverse-square law to simulate the brightness variation of an object according to its depth:
\begin{equation*}
     \mathbf{I} \approx \frac{1}{d^2} \approx s^2
\end{equation*}
where $\mathbf{I}$ is the light intensity of the object, $d$ is the depth and $s$ is the scale of the object blended in the image.

%% file: tex/04_MAV_dataset.tex
\subsection{MAV Dataset}
\begin{figure*}
    \centering
    \includegraphics[width=0.98\textwidth]{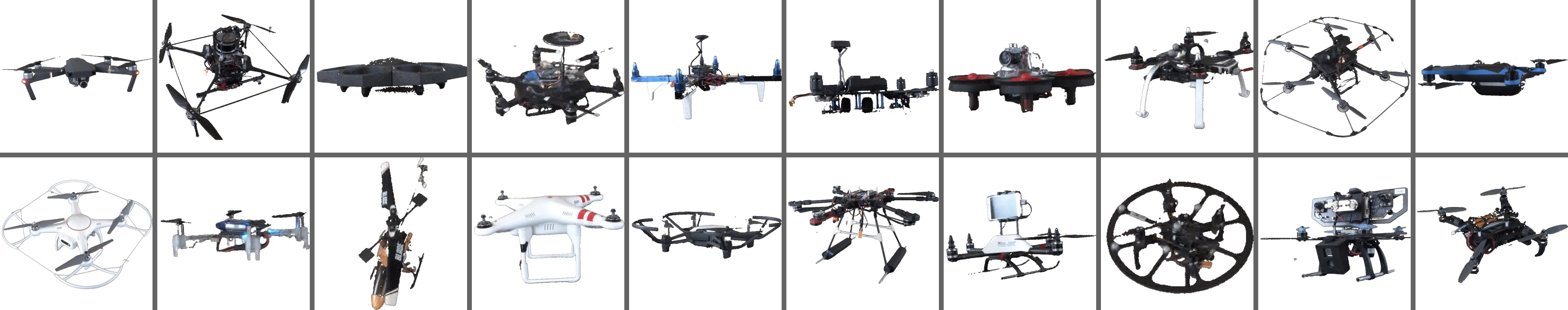}
    \caption{Example of MAV images in the PennSyn2Real dataset}
    \label{fig:multi_robot_grid}
\end{figure*}

We collect around $\SI{100,000}{}$ images of $21$ different MAVs including our prototypes and commercial drones such as \textit{Skydio2}, \textit{Mavic Pro}, \textit{Mavic Air}, and \textit{Autel X-Star}. Example of MAV images are shown in Fig.~\ref{fig:multi_robot_grid}. 

%% file: tex/05_experiments.tex
\section{Experiments}
In this section, we present experimental results using the synthetic data to train CNNs for MAV semantic segmentation and detection problems.


\subsection{Semantic Segmentation} 
\begin{figure}
    \centering
    \includegraphics[width=0.98\linewidth]{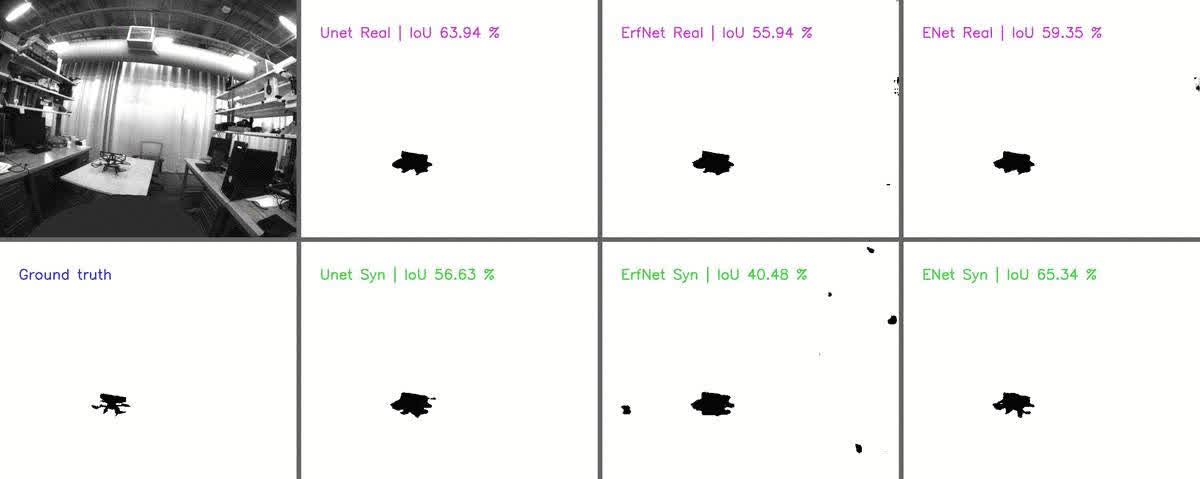}
    \caption{Qualitative performance comparison in semantic segmentation between models trained with real images (top row) and these trained with synthetic images (bottom row)}
    \label{fig:sementation_example}
\end{figure}

Saleh et al.~\cite{sadat2018effective} argue that objects such as people, cars, and vehicles are not suitable for syn-to-real transfer in segmentation problems due to the unrealistic characteristics of synthetic data. In this experiment, we show that our framework does not suffer from this problem. 
We conduct a semantic segmentation experiment using the test set provided by Nguyen et al.~\cite{nguyen2019mavnet} which features an FLA-250 drone.  To create a synthetic dataset,  we collect $12,000$ background images in the same room before superimposing FLA-250 images to create $10,000$ training images and $2000$ validation images. We train three different semantic segmentation models: UNet~\cite{ronneberger2015u}, ErfNet~\cite{romera2017erfnet}, and ENet~\cite{paszke2016enet} using the same set of hyperparameters as given in~\cite{nguyen2019mavnet}. Results provided in Tab.~\ref{tab:segmentation_performance} show that all models trained and validated with the synthetic data have competitive performance to their counterparts trained with real images. Fig.~\ref{fig:sementation_example} shows an example of the obtained results. 

\begin{table}[t]
\vspace{.2cm}
\centering
\begin{tabular}{|l ||c c c | c c c|}
\cline{1-7}
& \multicolumn{3}{c|}{Manually-Labeled Data} & \multicolumn{3}{c|}{Synthetic Data}  \\ \cline{2-7} 
    &UNet & ErfNet & ENet  & UNet & ErfNet & ENet \\ \hline \hline
IoU&     $63.50$& $53.50$& $67.62$& $61.4$& $41.62$& $63.62$\\
FN Rate& $14.27$& $1.97$&  $5.8$ &  $16.5$& $0.99$& $8.46$ \\
FP Rate& $0.14$&  $0.31$&  $0.16$ & $0.14$& $0.55$& $0.18$ \\
\cline{1-7}
\end{tabular}
\caption{Quantitative performance comparison between semantic segmentation models trained with synthetic data with those trained with real data with~\cite{nguyen2019mavnet} dataset.}
\label{tab:segmentation_performance}
\vspace{-3mm}
\end{table}

\begin{figure*}[t]
    \centering
    \includegraphics[width=0.98\textwidth]{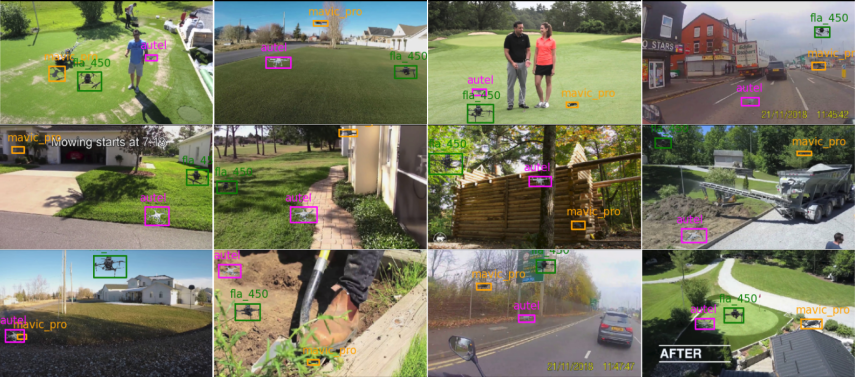}
    \caption{An example of synthetic outdoor multi-MAV data for training multi-object detection}
    \label{fig:syn_training_example}
\end{figure*}

\begin{table*}[h]
\vspace{.2cm}
\centering
\begin{tabular}{l ||c c c c | c c c c| c }
& \multicolumn{4}{c|}{Autel} & \multicolumn{4}{c|}{FLA-450} & mAP $\uparrow$\\ 
    Methods & AP $\uparrow$ & Precision $\uparrow$ & Recall  $\uparrow$ & F1  $\uparrow$ 
    & AP $\uparrow$ & Precision $\uparrow$ & Recall  $\uparrow$ & F1  $\uparrow$\\ \hline \hline
Zero-shot ImageNet  & $0.075$ & $0.002$ & $0.976$ & $0.005$ 
                    & $0.002$ & $0.002$ & $0.708$ & $0.005$ & $0.040$\\

1-shot ImageNet & $0.539$ & $0.013$ & $0.995$ & $0.026$ 
             & $0.003$ & $0.008$ & $0.250$ & $0.015$ & $0.271$\\   

5-shot ImageNet & $0.440$ & $0.027$ & $0.905$ & $0.052$ 
                & $0.164$ & $0.057$ & $0.619$ & $0.105$ & $0.302$\\   

50-shot ImageNet & $0.897$ & $0.239$ & $0.967$ & $0.383$ 
              & $0.268$ & $0.318$ & $0.300$ & $0.308$ & $0.58$\\
              
200-shot ImageNet & $0.977$ & $0.228$ & $1.000$ & $0.371$ 
               & $0.570$ & $0.863$ & $0.584$ & $0.696$ & $0.774$\\
Baseline (all-shot ImageNet) & $0.979$ & $0.303$ & $1.000$ & $0.464$ 
                             & $0.507$ & $0.782$ & $0.545$ & $0.642$ & $0.743$\\
\hline
Zero-shot Synthetic  & $0.939$ & $0.596$ & $0.948$ & $0.732$ 
               & $0.262$ & $0.798$ & $0.278$ & $0.412$ & $0.601$\\

1-shot Synthetic & $0.525$ & $0.090$ & $0.581$ & $0.156$ 
           & $0.693$ & $0.204$ & $0.778$ & $0.324$ & $0.610$\\   

5-shot Synthetic & $0.901$ & $0.274$ & $0.924$ & $0.423$ 
           & $0.651$ & $0.519$ & $0.715$ & $0.601$ & $0.776$\\   

50-shot Synthetic & $0.981$ & $0.309$ & $0.990$ & $0.471$ 
            & $0.626$ & $0.605$ & $0.702$ & $0.650$ & $0.803$\\
              
200-shot Synthetic & $0.940$ & $0.239$ & $0.971$ & $0.384$ 
             & $0.675$ & $0.448$ & $0.714$ & $0.550$ & $0.808$\\
              
All-shot Synthetic & $0.986$ & $0.296$ & $0.995$ & $0.456$ 
             & $0.651$ & $0.795$ & $0.684$ & $0.735$ & $\mathbf{0.819}$\\
\end{tabular}
\caption{Transfer learning using a model pretrained with the synthetic dataset in comparison with that model pretrained with ImageNet. ${N}-$prefix represents the number of real data samples for each object class such model uses during the finetuning. A model pretrained on the synthetic dataset performs better than its counterpart pretrained on ImageNet.}
\label{tab:quantitative_detection_performance}
\vspace{-3mm}
\end{table*}

\subsection{Object Detection}
\subsubsection{Dataset}
\textbf{Real Outdoor Multi-MAV:} 
We collect and manually label images from a video that captures an \textit{Autel X-Star}~\cite{autel} and an \textit{FLA-450} flying outdoors. We then split the data into two sets: a training set consisting of the first $1,200$ frames, and a test set consisting of the last $800$ frames. We split the training set further into a training set and a validation set with a $5:1$ ratio. 

\textbf{Synthetic Outdoor Multi-MAV:}
The corresponding synthetic dataset features object images of three MAVs: \textit{Autel X-star}, \textit{FLA-450}, and \textit{Mavic Pro}~\cite{mavic_pro}.  We collect background images of the same environment with that of the real outdoor multi-MAV data. We additionally download $8$ Youtube videos with keywords such as \textit{backyard}, and \textit{lawn} to diversify background images. In short, we generate roughly $31,000$ synthetic training samples and $14,000$ synthetic validation samples. Examples of these synthetic samples are shown in Fig.~\ref{fig:syn_training_example}.

\subsubsection{Baseline}
To provide a baseline for comparison, we finetune a Yolov3~\cite{redmon2018yolov3} model pretrained with ImageNet using the real training and validation sets. We use the Adam optimizer with the default learning rate and other parameters and train for $100$ epochs. We show the performance results in Tab.~\ref{tab:quantitative_detection_performance}. We also perform a similar test with the same Yolov3 model but do not finetune. This model's performance is reported under \textit{Zero-shot ImageNet} in the same table.  

\subsubsection{Synthetic to Real Transferability} 
\label{sec:transferablity}
In order to evaluate the transferability of a model trained with the synthetic data for object detection on the real multi-MAV test set, we first finetune the same Yolov3 model using the synthetic multi-MAV data with the same set of hyperparameters. Note that we do not use any real data here. We apply early stop after 15 epochs before performing a test on the real test set. Its result is shown as \textit{Zero-shot Synthetic} in Tab.~\ref{tab:quantitative_detection_performance}. As we can see, the zero-shot synthetic model's mAP is $0.601$ which is inferior to the baseline but much higher than that of the zero-shot ImageNet model. 

To further evaluate the transferability of a model trained with the synthetic data, we start with the zero-shot synthetic model that we finetune with the synthetic data above and provide an additional number of real samples to finetune more.  We randomly select from the real multi-MAV training set a subset that is guaranteed to contain $N$ object instances for every object class,  with $N \in {1, 5, 50, 200}$ to create corresponding $N$-shot training sets for few-shot learning.  

We finetune the zero-shot synthetic model using these subsets, one by one, with the same hyperparameters as before. After that, we test on the real-test set and report results in Tab.~\ref{tab:quantitative_detection_performance}. Note that during these finetuning processes, we also use the real multi-MAV validation set as we do with the baseline. Inspired by Nowruzi et al.~\cite{nowruzi2019much} which suggest that fine-tuning models trained on synthetic datasets with a small amount of real data is preferable to mixed training on a hybrid dataset with the same amount of real data, in these finetuning processes we do not reuse the synthetic data. All results are also shown in Tab.~\ref{tab:quantitative_detection_performance} as $N$-shot Synthetic, with $N \in {1, 5, 50, 200}$. 

To make a fair comparison, we also finetune a Yolov3 pretrained with ImageNet using those few-shot training sets, one by one, and provide the results in Tab.~\ref{tab:quantitative_detection_performance} as $N$-shot ImageNet, with $N \in {1, 5, 50, 200}$.

Interestingly, all $N$-shot Synthetic models perform better than their counterparts, with the $5$-shot Synthetic model already topping the baseline's performance. Fig.~\ref{fig:few-shot_detection_performance} highlights these observations. Furthermore, when providing the same amount of real data, all-shot synthetic - the model pretrained with the synthetic data outperforms the baseline which relies only on real data.





\begin{table}[h]
\vspace{.2cm}
\centering
\begin{tabular}{l ||c c c c}
& \multicolumn{4}{c}{Snapdragon}\\ 
    Methods & AP $\uparrow$ & Precision $\uparrow$ & Recall  $\uparrow$ & F1  $\uparrow$ \\ \hline \hline
Baseline & $0.343$ & $0.332$ & $0.378$ & $0.353$ \\

Zero-shot-model  & $0.274$ & $0.321$ & $0.366$ & $0.342$ \\

1-shot-model & $0.232$ & $0.100$ & $0.378$ & $0.154$ \\

5-shot-model & $0.236$ & $0.131$ & $0.377$ & $0.195$ \\

50-shot-model & $0.203$ & $0.191$ & $0.376$ & $0.254$ \\
              
200-shot-model & $0.161$ & $0.169$ & $0.376$ & $0.231$ \\
              
All-shot-model & $0.153$ & $0.252$ & $0.359$ & $0.296$ \\
\end{tabular}
\caption{Generalization on the snapdragon test set}
\label{tab:generalization_detection_performance}
\vspace{-3mm}
\end{table} 



\subsection{Generalization}
In this experiment, we investigate if a model trained with the synthetic dataset can generalize to detect other types of MAVs. To do so, we record and manually label a dataset that features snapdragon - a novel MAV that has not been seen during the training and finetuning of Yolov3 models in section~\ref{sec:transferablity}. We split these data into 3 parts: training with $1,200$ samples, validation with $200$ samples, and test with $800$ samples. Similar to the baseline in section~\ref{sec:transferablity}, we finetune a Yolov3 model pretrained with ImageNet using the snapdragon training set and report the performance on the test set in Tab.~\ref{tab:generalization_detection_performance} as \textit{baseline}. We then test all $N$-shot Synthetic models obtained from section~\ref{sec:transferablity} on this snapdragon test set without finetuning and report results in the same table. 
As can be seen, the zero-shot synthetic model has the closest performance to the baseline, $0.274$ in comparison with $0.343$. Furthermore, the degeneration in performance of $N$-shot synthetic model when $N$ increases from $zero$ to $all$ suggests that the more we finetune with the real multi-MAV real training set, the less generalization we can obtain.    

\begin{figure}
    \centering
    \includegraphics[width=0.9\linewidth]{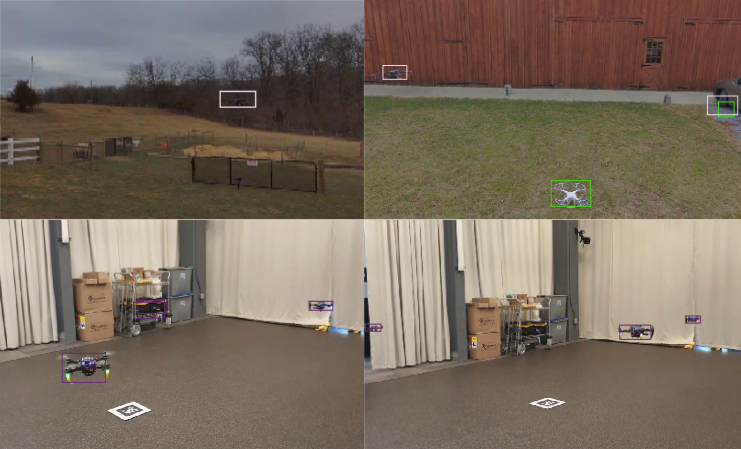}
    \caption{Examples of zero-shot synthetic model's performance on the real multi-MAV test set (top row) and the real snapdragon test set (bottom row)}
    \label{fig:detection_example}
    \vspace{-.5cm}
\end{figure}

%% file: tex/06_conclusions.tex
\section{Conclusions}
We introduce PennSyn2Real, a synthetic dataset with more than $100,000$ 4K images of different types of MAVs which can be used to generate an arbitrary number of labeled images for computer vision tasks. Our experiments demonstrate that the synthetic data generated using our framework can benefit a learning model for high-level computer vision tasks in all settings: training using only the synthetic data, and bootstrapping the synthetic data with the real data in few-shot learning. Additional research is necessary to improve the image blending process using other techniques such as generative models as well as making use of the object pose labels to generate synthetic data for object tracking or object pose optimization problems. Our ultimate goal is build on our preliminary work in \cite{NguyenICRA2020} to use PennSyn2Real to train detection and classification algorithms for drone-to-drone detection to enable formation flight in complex environments.